\documentclass[10pt, a4paper]{article}

\usepackage[]{lrec-coling2024} % this is the new style

\usepackage{graphicx}
\usepackage{amssymb}
\usepackage{amsmath,cases}
\usepackage{natbib}
\usepackage{multirow}
\usepackage{subfigure}
\usepackage{epstopdf}
\usepackage{comment}
\usepackage{url}
\usepackage{xcolor}
\usepackage{bbm}
\usepackage{color}
\usepackage{soul}
\usepackage{tcolorbox}
\usepackage{array}
\usepackage{enumitem}
\usepackage{arydshln}
\usepackage{booktabs}
\usepackage{multirow}
\usepackage{amsthm,amsmath,amssymb}
\usepackage{mathrsfs}
\usepackage{multirow}
\usepackage{setspace}
\usepackage{threeparttable}
\usepackage{tabu}
\usepackage{bm}
\usepackage{CJKutf8}

\tcbuselibrary{theorems}
\tcbset{highlight math/.append style={left=0mm,right=0mm,top=0mm,bottom=0mm, colframe=white}}
\newtcbox{\mybox}[1][red]
  {on line, arc = 0pt, outer arc = 0pt,
    colback = #1!10!white, colframe = #1!50!black,
    boxsep = 0pt, left = 1pt, right = 1pt, top = 2pt, bottom = 2pt,
    boxrule = 0pt}

\newtcbox{\wrongbox}[1][blue]
  {on line, arc = 0pt, outer arc = 0pt,
    colback = #1!10!white, colframe = #1!50!black,
    boxsep = 0pt, left = 1pt, right = 1pt, top = 2pt, bottom = 2pt,
    boxrule = 0pt}

\title{Chinese Sequence Labeling with Semi-Supervised Boundary-Aware Language Model Pre-training}

\name{
  \begin{tabular}{c}
    Longhui Zhang$^{1}$, Dingkun Long, Meishan Zhang$^{1}$\sthanks{\ \ Corresponding author.} \\
    Yanzhao Zhang, Pengjun Xie, Min Zhang$^{1}$\\
  \end{tabular}
}
\address{
    \begin{tabular}{c}
    $^{1}$ Harbin Institute of Technology (Shenzhen), Shenzhen, China, \\
        \{longhuizhang97,longdingkun1993,zhangyanzhao00,xpjandy\}@gmail.com \\
        \{zhangmeishan,zhangmin2021\}@hit.edu.cn
      \end{tabular}
}

\abstract{
Chinese sequence labeling tasks are heavily reliant on accurate word boundary demarcation. Although current pre-trained language models (PLMs) have achieved substantial gains on these tasks, they rarely explicitly incorporate boundary information into the modeling process.
An exception to this is BABERT~\cite{babert}, which incorporates unsupervised statistical boundary information into Chinese BERT's pre-training objectives. 
Building upon this approach, we input supervised high-quality boundary information to enhance BABERT's learning, developing a semi-supervised boundary-aware PLM.
To assess PLMs' ability to encode boundaries, we introduce a novel ``Boundary Information Metric'' that is both simple and effective. This metric allows comparison of different PLMs without task-specific fine-tuning. Experimental results on Chinese sequence labeling datasets demonstrate that the improved BABERT variant outperforms the vanilla version, not only on these tasks but also more broadly across a range of Chinese natural language understanding tasks. Additionally, our proposed metric offers a convenient and accurate means of evaluating PLMs' boundary awareness. 
\\ \newline \Keywords{Sequence Labeling, Boundary-Aware Language Model, Boundary Information Metric}}

\begin{document}
\begin{CJK}{UTF8}{gbsn}
\maketitleabstract
\section{Introduction}
Sequence labeling is an important task in Chinese natural language processing (NLP), encompassing various tasks such as Chinese word segmentation (CWS), part-of-speech (POS) tagging, and named entity recognition (NER). These tasks inevitably rely on boundary identification among various grained semantic units, which are unavailable from the input Chinese sentences~\cite{DBLP:conf/acl-sighan/Emerson05,jin-chen-2008-fourth}. There have been extensive studies that incorporate different types of boundary information into task-specific supervised machine learning models, i.e., subword-based models~\cite{yang-etal-2019-subword,li-etal-2021-char}, lexicon-enhanced models~\cite{zhang-yang-2018-chinese, diao-etal-2020-zen,lebert}. The results of these studies consistently demonstrate the high effectiveness of explicit boundary modeling in improving the performance of sequence labeling tasks in Chinese NLP.

Recently, Chinese PLMs, like BERT~\cite{DBLP:conf/naacl/DevlinCLT19}, have shown significant success in various NLP tasks~\cite{DBLP:conf/acl/WeiSWTC20,gao-etal-2021-simcse,DBLP:conf/naacl/ZhongC21}, including sequence labeling~\cite{DBLP:conf/acl/YangZD17,DBLP:conf/emnlp/JiangLSZXX21}. BERT is efficient in capturing general semantic information, but it overlooks boundary information required for Chinese sequence labeling~\cite{DBLP:conf/naacl/DevlinCLT19,DBLP:conf/emnlp/DiaoBSZW20,cui2021pre}.
Besides, substantial opportunities remain in research on the integration of boundary information into PLMs.
By encoding boundary information into PLMs, we can potentially improve the performance of PLMs on various Chinese NLP tasks without task-specific optimizations, greatly benefiting the Chinese NLP community.

Chinese BABERT~\cite{babert} is one of the few exceptions that inject unsupervised statistical boundary information into vanilla BERT, resulting in considerable performance gains on Chinese sequence labeling tasks. 
Nevertheless, BABERT has a notable limitation: due to the long tail problem in calculating these unsupervised statistical signals, the statistical boundary information extracted from raw mining corpus could be unstable and low-quality. As such, there is an opportunity to further improve performance by exploring alternative sources of higher-quality boundary information more closely aligned with human intuitions.

Along the line of BABERT, in this work, we present Semi-BABERT, which supplements supervised boundary signals to BABERT. We construct a large-scale lexicon from open sources, which serves as a reliable resource for high-quality boundary information. To enhance BABERT pre-training, we design a span-based boundary recognition objective based on the boundary information extracted from the lexicon. Considering that boundary identification from a lexicon may be incomplete, we propose the utilization of Positive-Unlabeled learning (PU learning)~\cite{DBLP:conf/ecml/LiL05, DBLP:conf/acl/PengXZFH19,DBLP:conf/aaai/HuL0JM0021} to address this limitation and enable auto-complementation. 
Additionally, we introduce a practical metric to quantify the potential of Chinese PLMs in encoding boundaries without task-specific fine-tuning,
which would be useful in swiftly assessing the adaptability of PLM to Chinese sequence labeling tasks.

\begin{figure*}[ht]
	\centering
	\includegraphics[width=0.7\linewidth]{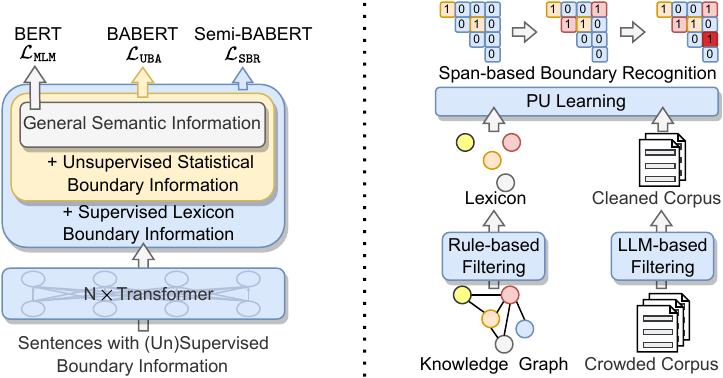}
	\caption{The relationship between BERT, BABERT and Semi-BABERT (left) and the supervised boundary recognition pre-training of Semi-BABERT (right).}
	\label{fig:model}
\end{figure*}

We conduct comprehensive experiments to evaluate Semi-BABERT's capability in Chinese sequence labeling, covering various CWS, POS, and NER datasets. The results consistently demonstrate that Semi-BABERT improves performance across all benchmark datasets. 
In an effort to further substantiate the model's efficacy, we broaden the scope of our evaluation to other Chinese natural language processing tasks, such as text classification and machine reading comprehension. Similarly, we observe a marked increase in performance in these tasks, reaffirming the effectiveness of Semi-BABERT in a series of Chinese language understanding tasks. Finally, we present an in-depth analysis specifically focused on Chinese sequence labeling tasks. This analysis offers valuable insights into the impact of Semi-BABERT on different aspects of sequence labeling\footnote{The source code and pre-trained Semi-BABERT will be publicly released at \url{http://github.com/modelscope/adaseq/examples/semibabert}.}.

\section{Method}

In this section, we first briefly introduce BABERT, which serves as the foundation for our work. We then present our Semi-BABERT, which leverages a lexicon to incorporate large-scale, high-quality supervised boundary information into BABERT. Finally, a novel ``Boundary Information Metric'' is proposed, which can swiftly and efficiently quantify the model boundary awareness without task-specific fine-tuning.

\subsection{BABERT}

As depicted on the left side of Figure~\ref{fig:model}, BABERT \cite{babert} incorporates unsupervised statistical boundary information into BERT's pre-training process, which consists of three steps.

1) The first step involves constructing a dictionary $\mathcal{N}$ to store the statistical boundary information. Each entry in the dictionary is a key-value pair, where the key represents an N-gram $g = \{c_1, \dots, c_m\}$ derived from the corpus. The corresponding value comprises three statistical boundary measures: Pointwise Mutual Information (\texttt{PMI}), Left Entropy (\texttt{LE}), and Right Entropy (\texttt{RE}).

2) Utilizing the dictionary $\mathcal{N}$, a boundary-aware representation $\mathbf{E} = \{\mathbf{e_1}, ..., \mathbf{e_n} \}$ is constructed for each text $x=\{c_1, \dots, c_n\}$ in the corpus. This representation incorporates the statistical boundary information.

3) The statistical boundary information is injected into the model using the unsupervised boundary-aware learning pre-training task (UBA task). Additionally, to capture general semantic information, BABERT incorporates BERT's Masked Language Modeling (MLM) task. Therefore, the overall loss function for BABERT is defined as follows:
\begin{equation}
	\label{eq:BABAERT}
	% \small
        \mathcal{L}_\texttt{BABERT} = \mathcal{L}_\texttt{MLM} + \mathcal{L}_\texttt{UBA}
\end{equation}
The above description provides a high-level overview of BABERT to facilitate understanding of our proposed method. For a more comprehensive understanding, one can refer to the original paper~\cite{babert}.

\subsection{Semi-BABERT}
To enhance the boundary encoding capability of PLM, we introduce Semi-BABERT, a novel approach that incorporates supervised lexicon boundary information into BABERT through a pre-training task called supervised boundary recognition (SBR task). As depicted on the right side of Figure~\ref{fig:model}, the training of Semi-BABERT consists of several modules: data source, data processing, and a new pre-training objective.

\paragraph{Data Source}
Our data is derived from two key sources: a knowledge graph and a crowded corpus for pre-training. The knowledge graph provides supervised lexical boundary information via rule-based filtering, whereas the crowded corpus supplies high-quality text data filtered by a large language model (LLM).
Specifically, we employ the OwnThink Knowledge Graph\footnote{\url{https://www.ownthink.com}}, which encompasses both entity and regular words, providing rich boundary cues. For the crowded corpus, in line with BABERT~\cite{babert}, we compile a mixed corpus from Chinese Wikipedia\footnote{\url{https://zh.wikipedia.org/wiki/}} and Baidu Baike\footnote{\url{https://baike.baidu.com/}} for our pre-training. This corpus consists of $3$ billion tokens and $62$ million sentences.

\paragraph{Data Preprocessing} Data preprocessing plays a crucial role in ensuring the quality of training data. Here we describe the two main techniques we employ for data preprocessing: rule-based lexicon filtering and LLM-based corpus filtering.

\begin{itemize}[label=\textcolor{black}{$\bullet$}, wide=0pt]
    \item  \emph{Rule-based Lexicon Filtering}: To ensure lexical quality, we implement rule-based lexicon filtering, applying constraints to remove undesirable words. Specifically, we limit the length of words to 2-4 characters, remove shorter words that are nested or overlapping with other words, and eliminate words that contain punctuation marks or English characters. After applying these filtering rules, we obtain a lexicon of $30$ million words.
    
    \item \emph{LLM-based Corpus Filtering}: The quality of the pre-training corpus is crucial for the performance of PLMs~\cite{DBLP:journals/corr/abs-1907-11692}. In our work, we leverage the power of LLMs for data cleaning to ensure the quality of the corpus used for pre-training. LLMs, thanks to their well pre-training and large-scale parameters, have demonstrated remarkable success in various natural language processing tasks \cite{DBLP:journals/corr/abs-2211-09110} and are known to generate high-quality text \cite{DBLP:journals/corr/abs-2304-00723}. Therefore, we rely on the powerful generation capabilities of LLMs for data cleaning.

    To evaluate the text quality in the crowded corpus, we introduce a task called ``Text Quality Evaluation''. The task prompt $\mathcal{P}$ is ``请生成一个语法规范、表达准确、逻辑严谨的中文文本'' (Please generate a grammatically correct, accurately expressed, and logically rigorous Chinese text). Given the task prompt $\mathcal{P}$ and a text $x=\{c_1, \dots, c_n\}$, we calculate the text quality score $s(\mathcal{P},x)$ using the following equation:
    \begin{equation}
    s(\mathcal{P},x) = \frac{1}{n} \sum_i \log p\left(c_i \mid c_{<i}, \mathcal{P}\right) \label{dataclean}
    \end{equation}
    $p\left(c_i \mid c_{<i}, \mathcal{P}\right)$ represents the generation probability of the character $c_i$ conditioned on the prompt $\mathcal{P}$ and the previous text $c_{<i}$. We use the Qwen-7B \cite{bai2023qwen} LLM for corpus filtering. To ensure efficiency, we first deduplicate the corpus and then evaluate the quality of all sentences in the corpus using Eq.~\ref{dataclean}. Finally, we remove the $10\%$ of the corpus with the lowest score, thereby retaining $90\%$.
\end{itemize}

\paragraph{PU Learning} PU learning algorithm~\cite{DBLP:conf/ecml/LiL05,DBLP:conf/acl/PengXZFH19} trains a binary classifier $f$ using only labeled positive examples $\mathcal{D}_p$ and a mixture of unlabeled positive and negative examples $\mathcal{D}_u$. This algorithm has shown success in distantly supervised NER~\cite{DBLP:conf/acl/PengXZFH19}. In PU learning, the loss function $\mathcal{L}(f)$ is defined based on ${\mathcal{D}_p}$ and ${\mathcal{D}_u}$ as follows:
\begin{equation}
    \begin{small}
            \mathcal{L}(f) \text{=} \gamma  \pi_p {\mathcal{L}}_p^{\text{+}}(f) \text{ + } \max \left\{0, {\mathcal{L}}_u^{\text{-}}(f) \text{ - } \pi_p {\mathcal{L}}_p^{\text{-}}(f)\right\} 
        \label{eq:pul}
     \end{small}
\end{equation}
where ${\mathcal{L}}_{p/u}^{\text{-}/\text{+}}(f)$ represents the loss of an example ${x} \in \mathcal{D}_{p/u}$ conditioned on an assumed positive (denoted as ``+'') or negative (denoted as ``-'') label. The function $\mathcal{L}$ is a binary cross entropy loss function. $\pi_p$ and $\gamma$ are the hyperparameters of PU learning. $\pi_p$ is the pre-estimated ratio of positive examples within $\mathcal{D}_u$. $\gamma$ is the loss weight of ${\mathcal{L}}_{p}^{\text{+}}(f)$. The paper~\cite{DBLP:conf/acl/PengXZFH19} proves a detailed proof of Eq.~\ref{eq:pul} in the PU learning algorithm.

%is the experience loss of example ${x} \in \mathcal{D}_{p/u}$ conditioned on positive (denoted as $1$) or negative (denoted as $0$) label. $\ell$ is a loss function, such as binary cross entropy. $n_{p/u}$ is the scale of set $\mathcal{D}_{p/u}$. $\pi_p$ is the ratio of positive examples within $\mathcal{D}^u$. A detailed proof of Eq.~\ref{eq:pul} is presented in~\citet{DBLP:conf/acl/PengXZFH19}.

\paragraph{Boundary Recognition Pre-training} To address the instability and low quality of statistical boundary information in BABERT, we propose Semi-BABERT, which incorporates supervised lexicon boundary information. We introduce a supervised boundary recognition (SBR) task that aims to identify word boundaries in text based on the lexicon.

Traditionally, boundary recognition is regarded as a sequence labeling problem \cite{DBLP:journals/ijclclp/Xu03, DBLP:conf/acl/DingLXZXWZ20}. However, this approach has two limitations. Firstly, it cannot handle nested boundaries, such as the distinction between ``南京'' (Nanjing) and ``南京市'' (Nanjing City). Secondly, supervision method heavily relies on lexical comprehensiveness, yet despite extensive size, the lexicon is unable to encompass all boundary information.

For the first drawback, we draw inspiration from research on structured information extraction \cite{DBLP:journals/corr/abs-2005-07150, DBLP:conf/emnlp/RenZYZLLL21} and adopt a span-based boundary recognition strategy instead of sequence labeling. This strategy involves training a binary classifier to determine whether N-grams in the text correspond to words. To address the second limitation, when dealing with an incomplete lexicon, we employ PU learning \cite{DBLP:conf/ecml/LiL05,DBLP:conf/acl/PengXZFH19,DBLP:conf/aaai/HuL0JM0021} to estimate the model loss under the ideal complete lexicon, and gradually expand the lexicon to the ideal completeness through multiple iterations. 

In our work, we focus on the span-based boundary recognition task (SBR). Given a lexicon and all N-grams present in the text, we consider the N-grams that exist in the lexicon as positive examples ${\mathcal{D}_p}$, while the remaining N-grams as unlabeled examples ${\mathcal{D}_u}$. To train a binary classifier $f$ with PU learning, we formalize the PU learning algorithm using Eq.~\ref{eq:pul}. During the training process, this algorithm helps us label the unlabeled N-grams in ${\mathcal{D}_u}$ as follows: If an unlabeled N-gram is predicted as a positive example by the classifier $f$ consecutively for $k$ times ($k$ is set to 5), we add the N-gram to the lexicon in the next iteration. The overall loss $\mathcal{L}_\texttt{SBR}$ of SBR task and the classifier $f$ are defined as follows:
\begin{equation}
	\begin{aligned}
    	\label{eq:wd}
            \mathcal{L}_\texttt{SBR} &= \mathcal{L}(f) \\
            f((\textbf{h}_i,\textbf{h}_j)) &= \texttt{sigmoid}(W\left[\textbf{h}_i;\textbf{h}_j\right]+b)
        \end{aligned}
\end{equation}
The span-based boundary recognition binary classifier $f$ is trained to determine whether the N-gram is a correct boundary. It takes the PLM representations ${\textbf{h}_i}$ and ${\textbf{h}_j}$ corresponding to the left and right boundary characters of the N-gram $\{c_i,\dots,c_j\}$ as input. The classifier predicts the probability that the N-gram is a word, using a sigmoid activation function applied to the linear transformation of the concatenation of ${\textbf{h}_i}$ and ${\textbf{h}_j}$ with weight matrix $W$ and bias $b$.

%where ${h_i}$ and ${h_j}$ are the BERT representations corresponding to the left and right boundary characters of the N-gram $\{c_i,\dots,c_j\}$. $p_{i,j}$ is the probability that classifier $f$ predicts that N-gram is a word, and $y_{i,j}$ is the label assumed by the PU learning strategy in Eq.~\ref{eq:pul} for N-gram.
\begin{figure}
	\centering
	\includegraphics[width=1.0\linewidth]{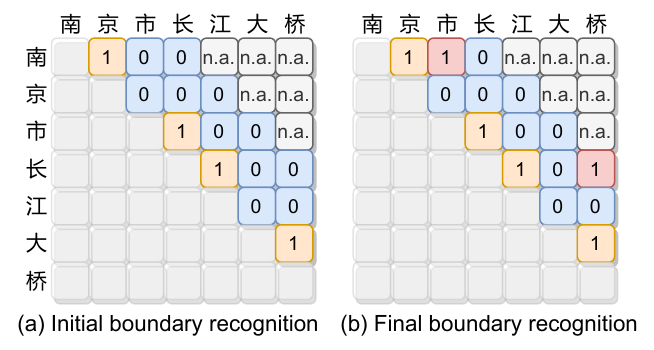}
	\caption{Boundary identification results in the initial and final stages of the SBR task.}
	\label{fig:bd}
\end{figure}
In Figure~\ref{fig:bd}, we demonstrate an example of the SBR task with PU learning. We consider the text ``南京市长江大桥'' (Nanjing Yangtze River Bridge). 
In the initial stage, only four words ``南京'' (Nanjing), ``市长'' (mayor), ``长江'' (Yangtze River), and ``大桥'' (Bridge) are included in the lexicon, as shown in Figure~\ref{fig:bd}(a). However, with the help of PU learning and multiple iterations of the model, the boundary information of all N-grams such as ``南京市'' (Nanjing City) and ``长江大桥'' (Yangtze River Bridge) can be identified and added to the lexicon, as illustrated in Figure~\ref{fig:bd}(b). It is important to note that we only consider N-grams $\{c_i,\dots,c_j\}$ where $1<j-i<4$. Hence, the lower triangle area is ignored as $j-i \leq 1$, and certain parts of the upper triangle area do not require prediction as $j-i \geq 4$.

\paragraph{Pre-training Objective} 
The architecture of Semi-BABERT, depicted on the left side of Figure~\ref{fig:model}, builds upon the foundations of BERT and BABERT. Therefore, apart from the span-based boundary recognition task, Semi-BABERT aligns with the training objectives of these base models. Consequently, the total pre-training loss for Semi-BABERT can be formulated as follows:
\begin{equation}
\mathcal{L}_\texttt{Semi-BABERT}=\mathcal{L}_\texttt{MLM}+\mathcal{L}_\texttt{UBA}+\mathcal{L}_\texttt{SBR}
\end{equation}

\subsection{Boundary Information Metric}
Given the close relationship between the boundary awareness capability of PLMs and their performance on downstream Chinese sequence labeling tasks during fine-tuning, it is crucial to evaluate the boundary awareness capability of these PLMs. Previous research relied on downstream task performance as an evaluation metric~\cite{lebert,babert}. Although this evaluation metric is valid, it is unintuitive and resource-consuming. To address this issue and provide a more reasonable assessment for Chinese PLMs’ boundary awareness, we introduce a novel evaluation metric called the “Boundary Information Metric” (BIM). This metric can evaluate PLMs' boundary recognition ability without task-specific fine-tuning.

We begin by assuming that: an ideal boundary-aware sentence representation should exhibit higher similarity between characters within words~\footnote{This assumption is empirically verified in Section~\ref{exbim}}. To quantify the boundary information, we consider a character $c$ and define any other character within the same word as a positive sample $c^+$, while the character in a different word is considered to be a negative sample $c^-$. We calculate the similarity between $c$ and $c^+$, denoted as $\texttt{SIM}_{pos}$. Similarly, the similarity between negative samples $c$ and $c^-$ is denoted as $\texttt{SIM}_{neg}$. The BIM is then calculated based on the difference between $\texttt{SIM}_{pos}$ and $\texttt{SIM}_{neg}$. We employ the cosine function to calculate the similarity between the vector representations $\textbf{h}$, $\textbf{h}^+$, and $\textbf{h}^-$ corresponding to $c$, $c^+$, and $c^-$, respectively. 
A higher BIM indicates a stronger boundary awareness in the model. The final computation process of BIM can be formulated as:
\begin{equation}
	\begin{aligned}
		\texttt{SIM}_{pos} &=  \underset{(\textbf{h},\textbf{h}^+) \sim p_{\mathrm{pos}}}{\mathbb{E}}  \texttt{sim} \left(\textbf{h}, \textbf{h}^+\right) \\
		\texttt{SIM}_{neg} &=  \underset{(\textbf{h},\textbf{h}^-) \sim p_{\mathrm{neg}}}{\mathbb{E}}  \texttt{sim} \left(\textbf{h}, \textbf{h}^-\right) \\
		\texttt{BIM} &= \texttt{SIM}_{pos} -\texttt{SIM}_{neg}
		\label{eq:bii}
	\end{aligned}
\end{equation}

Intuitively, the similarity between character pairs not only depends on whether they are within the same word but also on their distance from the whole text. Characters farther apart tend to have lower similarity. To mitigate the influence of character distance on the BIM, we constrain the distance between the negative sample pair ($c$, $c^-$). Specifically, we make DIS($c$, $c^-$) approximately equal to DIS($c$, $c^+$), where DIS notes the distance between characters. Since $c$ and $c^+$ are within the same word and are typically close to each other, we restrict the distance between $c$ and $c^-$, i.e., the DIS($c$, $c^-$) to a value less than a threshold $L$. In our works, we set $L$ to $2$. This constraint facilitates accurately quantify the boundary awareness of PLMs regardless of the character distance in the text.

\setlength{\arrayrulewidth}{0.6pt}
\begin{table*}[ht]
	\centering
	\small
	\resizebox{\textwidth}{26mm}{
		\begin{tabular}{ccccccccccccccc}
			\toprule
			\multirow{2}{*}{\bf Model} & \multicolumn{3}{c}{\bf CWS} & \multicolumn{3}{c}{\bf POS} & \multicolumn{7}{c}{\bf NER} & {\bf AVG} \\
			\cmidrule(r){2-4} \cmidrule(r){5-7} \cmidrule(r){8-14}
			& CTB6       & MSRA       & PKU        & CTB6       & UD1        & UD2        & Onto4      & Book       & News       & Finance  &MSRA &Resume	&Weibo & {\bf Score}   \\ \midrule
			BERT-wwm          &97.4 &98.3 &96.5 &94.8 &95.5 &95.4 &80.9 &76.2 &79.3 &85.0 &95.7 &95.8 &68.6 &89.2   \\ 
			ERNIE-Baidu         &97.4 &98.2 &96.3 &94.9 &95.3 &95.1 &80.4 &76.6 &80.4 &86.0 &95.1 &95.6 &70.0 &89.3   \\ 
			ERNIE-Gram          &97.3 &98.3 &96.4 &94.9 &95.3 &95.2 &81.0 &77.2 &80.0 &85.3 &95.8 &95.6 &68.4 &89.3  \\
			ZEN          &97.3 &98.3 &96.5 &94.8 &95.5 &95.5 &80.1 &75.7 &80.2 &85.0 &95.2 &95.4 &66.7 &88.9    \\
			NEZHA         &\bf 97.5 &\bf 98.6 &96.7 &95.0 &95.6 &95.5 &81.7 &77.0 &79.8 &85.2  &\bf 96.6 &95.7 &70.3     &89.6   \\ 
			\hdashline
   BERT          &97.3 &98.2 &96.3 &94.7 &95.0 &94.9 &81.0 &76.1 &79.2 &85.3 &95.8 &95.6 &69.6 &89.2    \\
   BABERT     &97.5 &98.4 &96.7 &95.0 &95.7 &95.5 &81.9 &76.8 &80.3 &86.9 &96.3 &95.8 &68.3 &89.6    \\
			Ours   &97.4 &\bf 98.6  &\bf{96.8}  &\bf{95.2} &\bf{95.7} &\bf{95.6}  & \bf{82.2} & \bf{80.7}  & \bf{81.9}    & \bf{87.1}  &96.3 &\bf{96.0} &\bf{71.0}   &\bf{90.3} \\ \midrule 
   BERT-lite &97.0 &98.1 &96.1 &94.4 &93.7 &93.3 &77.2 &76.0 &79.1 &83.9 &93.9 &95.4 &64.4 &87.9 \\ 
   Ours-lite &97.0 &98.2 &96.4 &94.5 &94.6 &94.4 &78.8 &80.1 &80.2 &84.3 &94.4 &95.5 &65.9 &88.8 \\   \midrule
   ChatGPT &94.3 &92.9 &93.1 &88.6 &92.0 &92.1 &69.4 &70.9 &80.4 &79.3 &90.1  &95.7 &70.1 &85.3 \\
\bottomrule
	\end{tabular}}
	\caption{Fine-tuning results on Chinese sequence labeling tasks. We report the F1-score on the test set.}
	\label{tab:Semi-BABERT} 
\end{table*}
\section{Experiments}
\subsection{Datasets}
In order to assess the effectiveness of Semi-BABERT for Chinese sequence labeling tasks, we conduct an extensive evaluation on a total of $13$ diverse datasets. These datasets encompass a range of Chinese NLP tasks, including CWS, POS, and NER. These tasks are particularly relevant for evaluating the boundary awareness of PLMs, as they heavily rely on accurate boundary information.

For the CWS task, we utilize three datasets: CTB6~\cite{xue2005penn}, MSRA, and PKU~\cite{emerson-2005-second}. For the POS task, we deploy three different datasets: CTB6~\cite{xue2005penn}, UD1, and UD2~\cite{nivre2016universal,shao-etal-2017-character}. Lastly, for NER, we employ a total of seven datasets: Onto4~\cite{weischedel2011ontonotes}, Book~\cite{jia-etal-2020-entity}, News~\cite{jia-etal-2020-entity}, Finance~\cite{jia-etal-2020-entity}, MSRA~\cite{levow-2006-third}, Resume~\cite{DBLP:conf/naacl/YangZL19}, and Weibo~\cite{DBLP:conf/emnlp/PengD15,DBLP:conf/acl/PengD16}.

\subsection{Experimental Settings}
We conduct pre-training of Semi-BABERT on a distributed setting using eight NVIDIA Tesla V100 GPUs, each equipped with 32GB memory. The hyperparameters and configurations of the baseline PLMs and Semi-BABERT are as follows:

\paragraph{Hyperparameters} Following BABERT~\cite{babert}, during pre-training, we use vanilla BERT to initialize the weights of Semi-BABERT. To accommodate different contexts, we pre-train two variations of Semi-BABERT, namely Semi-BABERT-base and Semi-BABERT-lite. Semi-BABERT-lite consists of 6 transformer layers, 8 self-attention heads, a hidden dimension of 512, and a total of 30 million parameters. In contrast, Semi-BABERT-base comprises 12 transformer layers, 12 self-attention heads, a hidden dimension of 768, and a total of 110 million parameters.
For both variations, we adopt the same training hyperparameters: a batch size of 4096, a learning rate of 1e-4, a warmup ratio of 0.1, a maximum sentence length of 512, and a maximum N-gram length of 4. In the PU learning equation (Eq.~\ref{eq:pul}), we set $\pi_p$ to 0.2 and $\gamma$ to 0.5.

\paragraph{Baselines} To evaluate the effectiveness of Semi-BABERT, we compare its performance against several baseline models, including the following state-of-the-art approaches: BERT-lite~\cite{DBLP:conf/naacl/DevlinCLT19}, BERT~\cite{DBLP:conf/naacl/DevlinCLT19}, BERT-wwm~\cite{cui2021pre}, ERNIE-Baidu~\cite{sun2019ernie}, ERNIE-Gram~\cite{xiao-etal-2021-ernie}, ZEN~\cite{diao-etal-2020-zen}, NEZHA~\cite{wei2019nezha}, and BABERT~\cite{babert}.

To ensure a fair and consistent comparison, we adopt the fine-tuning approach proposed by BABERT~\cite{babert}, which incorporates PLMs with a conditional random field (CRF) layer for sequence labeling. During the inference stage, we utilize the Viterbi algorithm to generate the optimal label sequence. To mitigate the effects of randomness, we perform fine-tuning using five different random seeds and subsequently average the results. This process allows us to obtain robust and reliable performance estimates for Semi-BABERT and the baselines, enabling a comprehensive and unbiased comparison.

\subsection{Main Results}
In this section, we present the fine-tuning results of Semi-BABERT on $13$ sequence labeling tasks, comparing it to various baselines. Table~\ref{tab:Semi-BABERT} summarizes the results, and we draw the following observations:

(1) Effectiveness of Semi-BABERT: Among models of the same scale, Semi-BABERT consistently achieves the highest average performance across all datasets. Specifically, Semi-BABERT-lite outperforms BERT-lite. When compared to the state-of-the-art BABERT, Semi-BABERT-base exhibits an average score increase of 0.8 (90.4-89.6) across all datasets. These results clearly demonstrate that Semi-BABERT surpasses PLMs of the same size, emphasizing the significance of supervised boundary information in achieving superior performance.

(2) Importance of boundary information: Apart from the scale of the training data, boundary information plays a crucial role in sequence labeling tasks. NEZHA, for instance, was pre-trained on three datasets (Chinese Wikipedia, Baidu Baike, and Chinese News) with a collective token count of 11 billion, four times larger than our dataset~\cite{wei2019nezha}. Compared to NEZHA, the average score of Semi-BABERT with enhanced boundary information increases by 0.7 (90.3-89.6). Furthermore, despite the limited training data from Resume and Weibo~\cite{lebert}, Semi-BABERT still outperforms other models. These findings indicate that the inclusion of boundary information can compensate for the lack of training data.

(3) Importance of model size: The incorporation of high-quality boundary information enables the 6-layer Semi-BABERT-lite to surpass the 12-layer BERT on the Book and News datasets of NER task. However, BERT still maintains an average score 0.4 points higher than Semi-BABERT-lite (89.2-88.8), underscoring the continued significance of the size of PLMs, despite the gains made by Semi-BABERT-lite from incorporating boundary information.

\begin{table}[t]
\centering
\resizebox{0.45\textwidth}{!}{
\begin{tabular}{p{6mm}p{70mm}}
\toprule
\multirow{2}{*}{CWS} & 请对文本进行分词。输出形式为``单词1/单词2''。 文本: [TEXT]。 输出: \\
 & Please segment the text into words. Output format is ``word1/word2''. Text: [TEXT]. Output:  \\ \midrule
 \multirow{2}{*}{POS} & 请对下面的文本进行词性标注，其中词性列表为[Category List]。输出形式为``type1: word1; type2: word''。 文本: [TEXT]。 输出:  \\
 & Please provide part-of-speech tagging for the following text, where the tag list is [Category List]. Output format is ``type1: word1; type2: word''. Text: [TEXT]. Output: \\ \midrule
 \multirow{2}{*}{NER} & 请列出文本中所有符合下列类别的实体。输出形式: ``类别1: 实体1; 类别2: 实体2;''。类别: [Category List] 文本: [TEXT]。 输出:   \\
 & Please list all entities in the text that fit the following category. Output format is ``type1: entity1; type2: entity2;''. Category: [Category List]. Text: [TEXT]. Output: \\
 \bottomrule
\end{tabular}}
\caption{ChatGPT prompts for three tasks.}
\label{tab:chatgpt} 
\end{table}

In addition to the Chinese PLMs, LLMs have also achieved impressive performance in various tasks~\cite{DBLP:journals/corr/abs-2211-09110}. ChatGPT~\footnote{\url{https://openai.com/blog/chatgpt}} is the most representative among them. To evaluate its performance on Chinese sequence labeling tasks, we conduct further tests. Due to ChatGPT's closed-source setup and large-scale parameters, fine-tuning it becomes challenging. As a result, we only test ChatGPT in unsupervised scenarios. The prompts used for evaluation are provided in Table~\ref{tab:chatgpt}.
The averages scores in Table~\ref{tab:Semi-BABERT} demonstrate that unsupervised ChatGPT remains less effective than supervised PLMs. Nevertheless, unsupervised ChatGPT exhibited noteworthy progress on certain datasets. Specifically, unsupervised ChatGPT outperforms supervised BERT on the News and Weibo datasets of the NER task. This outcome may be attributable to the limited scale of these datasets~\cite{lebert}, which likely provided inadequate training for the supervised PLMs.

\subsection{Analysis}

In this section, we provide a comprehensive analysis of Semi-BABERT from four different perspectives: few-shot setting, probing, BIM evaluation, and case study.

\begin{table}[t]
	\centering
            \resizebox{0.47\textwidth}{!}{
		\begin{tabular}{@{}lcccccc@{}}
			\toprule
			& \multicolumn{3}{c}{PKU}         & \multicolumn{3}{c}{Onto4}                     \\
			& 10             & 50             & 100         & 10             & 50             & 100            \\ \midrule
			BERT-wwm   &84.7 &88.0 &88.8 &12.8 &43.1 &59.4          \\
			ERNIE      &84.3 &87.0 &88.2 &19.9 &43.0 &50.8          \\
			ERNIE-Gram &84.0 &86.6 &88.0 &28.4 &45.9 &60.0          \\
			NEZHA     &84.4 &88.7 &89.7 &14.5 &44.1 &59.2          \\  \hdashline
                BERT       &84.0 &87.9 &88.2 &14.9 &42.4 &58.0          \\
                BABERT & 84.7 & 89.5 & 90.0 &32.1     & 46.6     & {60.6} \\
			Ours     &\bf 85.2 &\bf 90.5 &\bf 92.0 &\bf 50.8 &\bf 59.2 &\bf 64.5 \\  \midrule
                BERT-lite &80.3 &84.2 &85.6 &12.8 &40.1 &57.0 \\
			Ours-lite     &{84.8} &{89.6} &{91.8} &{47.6} &{57.1} &59.7 \\ \bottomrule
		\end{tabular}}
	\caption{Few-shot results on PKU (CWS task) and Onto4 (NER task), using 10, 50, and 100 examples of the training data.}
	\label{tab:few-shot}
\end{table}

 \paragraph{{Few-Shot}} PLMs have shown great performance in low-resource scenarios due to their extensive pre-training. To further investigate the capabilities of Semi-BABERT, we conduct fine-tuning experiments on various PLMs using $10$, $50$, and $100$ randomly selected examples from the original training data of PKU (CWS) and Onto4 (NER) datasets.

Table~\ref{tab:few-shot} presents the results of these few-shot experiments, and it is evident that Semi-BABERT-base consistently outperforms various baselines across all few-shot settings. In the 10-shot setting of Onto4, Semi-BABERT-base achieves a remarkable score increase of 18.7 (50.8-32.1) points compared to BABERT, showcasing its effectiveness in low-resource scenarios. Additionally, the 6-layer Semi-BABERT-lite performs better than any other 12-layer PLMs, except in the 100-shot scenario of Onto4. These findings highlight the considerable performance gains attainable by effectively encoding boundary information, particularly in few-shot situations.

\begin{table}[t]
        \centering
	\small
	\begin{tabular}{cccc}
		\toprule
		&PKU  &UD2 &Onto4 \\ \hline
            
            BERT-wwm    &88.8  &55.3  &32.5  \\
            ERNIE-Baidu   &88.4  &55.7  &42.2  \\
            ERNIE-Gram   &87.9  &54.9  &32.9  \\
            ZEN    &88.7  &55.6  &31.8  \\
            NEZHA  &88.7  &55.9  & 38.3  \\
             \hdashline
            BERT  &87.4  &54.7  &31.2  \\
            BABERT   &88.9  &56.1  &44.2  \\
            Ours   &\bf 89.2  &\bf 56.5  &\bf 45.7   \\ 
\bottomrule
	\end{tabular}
	\caption{The results of fine-tuning with frozen PLM on PKU (CWS), UD2 (POS) and Onto4 (NER).}
	\label{tab:additional}
\end{table}

\paragraph{{Probing}} The evaluation method based on full-parameter fine-tuning primarily assesses the performance of the fine-tuned models rather than the pre-trained ones. To gain further insights, we adopt a straightforward approach where the PLMs are frozen during fine-tuning, and only additional parameters, such as the CRF layer, are trained. Table~\ref{tab:additional} presents the results obtained using this fine-tuning method for various PLMs.

Interestingly, under the PLM frozen setting, Semi-BABERT demonstrates clear advantages over the full-parameter fine-tuning approach. Specifically, when comparing to BABERT in the full-parameter fine-tuning scenario (Table~\ref{tab:Semi-BABERT}), Semi-BABERT achieves improvements of 0.1, 0.1, and 0.3 on the PKU, UD2, and Onto4 datasets, respectively. However, in the PLM frozen scenario (Table~\ref{tab:additional}), these improvements are enhanced to 0.3, 0.4, and 1.5, respectively. This significant difference in performance highlights the effectiveness of Semi-BABERT's ability to enhance boundary awareness during pre-training.

\begin{table}[t]
        \centering
	\small
	\begin{tabular}{llll}
		\toprule
		&$\texttt{SIM}_{pos}$  &$\texttt{SIM}_{neg}$ &$\texttt{BIM}$ \\ \hline
		
		BERT-wwm   &72.0 &60.7 &11.2    \\
		ERNIE-Baidu &78.1 &64.5 &13.6  \\
		ERNIE-Gram &86.2 &72.6 &13.6  \\
            ZEN &81.7 &70.6 &11.1 \\
            NEZHA &48.9 &35.1 &13.8 \\
		 \hdashline
   BERT  &67.6 &57.1 &10.5   \\
   BABERT     &65.2 &51.2 &14.0           \\
		Ours  &62.5  &47.3 &\bf 15.2   \\ 
  \midrule
            BERT-lite &59.0 &48.6 &10.5 \\
		Ours-lite  &57.8 &43.6 &14.1       \\ 
\bottomrule
	\end{tabular}
	\caption{$\texttt{SIM}_{pos}$, $\texttt{SIM}_{neg}$ and $\texttt{BIM}$ of various PLMs.}
	\label{tab:bii}
\end{table}

\paragraph{{BIM Evaluation}}\label{exbim} In this subsection, we evaluate the effectiveness of the BIM, a boundary-aware quantification method that does not require task-specific fine-tuning. To assess the performance of BIM, we apply it to various PLMs. As BIM requires sentence segmentation, we conduct experiments on the CTB6 test set~\cite{xue2005penn} of the Chinese Word Segmentation task, which provides high-quality word segmentation annotations.

The results, presented in Table~\ref{tab:bii}, demonstrate that for all models, $\texttt{SIM}_{pos}$ (the similarity between characters within words) is consistently higher than $\texttt{SIM}_{neg}$ (the similarity between characters across word boundaries). This observation validates BIM's underlying assumption that character representations within words tend to be more similar. Additionally, Semi-BABERT-base consistently achieves a higher BIM score compared to other PLMs, reaffirming its superior boundary awareness.

Two phenomena are surprising: (1) BERT-lite and BERT yield the same BIM score. (2) Semi-BABERT-lite achieves a higher BIM score than the larger-scale BABERT. The results suggest that the BIM is independent of the model scale, as it primarily measures boundary awareness of the model. Consequently, BIM serves as a more accurate quantification method for boundary information than task-specific fine-tuning.

\begin{table}[]
\centering
\resizebox{0.5\textwidth}{!}{
\begin{tabular}{p{10mm}p{70mm}}
        \toprule
            \multirow{2}{*}{Text} 
            & \mybox{黑手党2}$_{\texttt{game}}$类似于游戏\mybox{《侠盗飞车》}$_{\texttt{game}}$，气氛有之前上映的\mybox{黑帮传奇}$_{\texttt{movie}}$的味道。 \\
            & \mybox{Mafia 2}$_{\texttt{game}}$ is akin to \mybox{Grand Theft Auto}$_{\texttt{game}}$ with an ambiance reminiscent of the prior movie \mybox{Gangster Legend}$_{\texttt{movie}}$. \\  \midrule
            BERT &  \wrongbox{黑手党}$_{\texttt{game}}$2类似于游戏\mybox{《侠盗飞车》}$_{\texttt{game}}$，气氛有之前上映的\wrongbox{黑帮}$_{\texttt{org}}$传奇的味道。 \\ 
            BABERT &  \mybox{黑手党2}$_{\texttt{game}}$类似于游戏\mybox{《侠盗飞车》}$_{\texttt{game}}$，气氛有之前上映的\wrongbox{黑帮}$_{\texttt{org}}$传奇的味道。 \\ 
            Ours &  \mybox{黑手党2}$_{\texttt{game}}$类似于游戏\mybox{《侠盗飞车》}$_{\texttt{game}}$，气氛有之前上映的\mybox{黑帮传奇}$_{\texttt{movie}}$的味道。 \\ \bottomrule
        \end{tabular}
    }
    \caption{Case study on NER task. \mybox{Red} (\wrongbox{Blue}) represents correct (incorrect) entities.}
    \label{tab:case-study}
\end{table}

\begin{table*}[ht]
	\centering
        \resizebox{\textwidth}{28mm}{
		\begin{tabular}{ccccccccccc}
			\toprule
			\multirow{2}{*}{\bf Model} & \multicolumn{6}{c}{\bf TC} & \multicolumn{3}{c}{\bf MRC} & {\bf AVG}\\
			\cmidrule(r){2-7} \cmidrule(r){8-10} 
			& AFQMC & TNEWS  & IFLYTEK & OCNLI & WSC & CSL & CMRC & ChID & C3  & {\bf Score} \\
			\midrule
			MacBERT &69.9 &57.9 &60.4 &67.4 &74.7 &82.1 &73.5 &79.5 &58.9 &69.4 \\
			PERT &73.6 &54.5 &57.4 &66.7 &76.1 &\bf 82.8 & 73.8 & 80.2 & 58.0 & 69.2 \\
			NEZHA &\bf 73.5 &58.5 &55.7 &\bf 69.0 &76.7 &82.6 &71.9 & 87.1 &{75.2}	 &72.2 \\ \hdashline
   ERNIE-THU & 72.9& 	56.6 & 	59.3& 	68.0& 	75.8& 	82.4& 	73.0& 	80.2& 	56.3 &69.4 \\ 
           ERNIE-Baidu &73.1 &56.2 &60.1 &67.5 &75.8 &82.1 &72.9 &80.0 &57.6 &69.5 \\
            K-BERT & 73.2& 	55.9& 	60.2& 	67.8& 	76.2& 	82.2& 	72.7& 	80.3& 	57.5 & 69.6\\
            CKBERT & 73.2 & 56.4 &60.7 &68.5 &76.4 &82.6 & 73.6 &81.7 & 57.9 &70.1 \\  \hdashline
            BERT &72.7 &55.2 &59.5 &66.5 &72.5 &81.8 &73.4 &79.2 &57.9 &68.8  \\
            BABERT &71.1 &57.1 &60.0 &65.6 &73.1 &80.2 &71.3 &83.1 &68.1 &70.0 \\ 
			Ours & 73.1 & \bf 59.4 & \bf 61.9 & 68.5 & \bf 80.3 &81.1 & \bf 73.9 &\bf {87.4} &  \textbf{77.0} & \bf 73.6 \\ 
			\bottomrule
		\end{tabular}}
	\caption{Performance of different PLMs on CLUE benchmarks.}
	\label{general_result}
\end{table*}

\paragraph{{Case Study}} To illustrate the clear advantages of Semi-BABERT over BERT and BABERT, we present a case study focusing on the NER task, as shown in Table~\ref{tab:case-study}. In this case, BERT, lacking explicit boundary information, fails to correctly identify the entities ``黑手党2'' (Mafia 2) and ``黑帮传奇'' (Gangster Legend). Although BABERT improves upon BERT by successfully identifying ``黑手党2'' (Mafia 2), it still falls short with an incomplete entity. Only Semi-BABERT accurately identifies all entities and their respective categories. This case study serves as a compelling demonstration of how high-quality boundary information can significantly enhance the performance of models on sequence labeling tasks.

\subsection{Additional Results on CLUE}

To assess the broader effectiveness of Semi-BABERT across various tasks, we conduct experiments on the widely-used CLUE benchmark~\cite{clue}, specifically focusing on Chinese text classification (TC) and machine reading comprehension (MRC) tasks. The TC task consists of 6 datasets: AFQMC, TNEWS, IFLYTEK, OCNLI, WSC, and CSL. The MRC task comprises 3 datasets: CMRC, ChID, and C3. Following~\cite{DBLP:journals/corr/abs-2210-05287}, we use the fine-tuning code provided by CLUE benchmarks\footnote{\url{https://github.com/CLUEbenchmark/CLUE/}}. 

In addition to boundary information, the CLUE benchmarks also require models to possess significant knowledge, including structured relational information in a knowledge graph~\cite{DBLP:journals/corr/abs-2210-05287}. Therefore, we compare Semi-BABERT with several knowledge-enhanced PLMs, namely ERNIE-THU~\cite{zhang-etal-2019-ernie}, ERNIE-Baidu~\cite{sun2019ernie}, K-BERT~\cite{DBLP:conf/aaai/LiuZ0WJD020}, and CKBERT~\cite{DBLP:journals/corr/abs-2210-05287}.

Table~\ref{general_result} presents the results of various PLMs on the CLUE benchmark. From an average score perspective, Semi-BABERT outperforms all PLMs, highlighting its significant advantage. Given the complexity of the CLUE benchmarks, knowledge-enhanced models typically exhibit superior performance compared to models like BERT and MacBERT. However, Semi-BABERT achieves exceptional results on this benchmark, even without additional knowledge information. Notably, Semi-BABERT surpasses the knowledge-enhanced CKBERT (70.1) and NEZHA (72.2), despite the latter models being pre-trained on larger corpus. This observation suggests that the inclusion of boundary information can compensate for the absence of extensive knowledge, demonstrating its compensatory effect on model performance.

\section{Related Work}
PLMs learn sentence representations through pre-trained tasks on a large-scale corpus. For example, BERT~\cite{DBLP:conf/naacl/DevlinCLT19} proposes two pre-trained tasks, Masked Language Modeling (MLM) and Next Sentence Prediction (NSP), to learn bidirectional sentence representation. Recent studies have explored extensions of the BERT model. RoBERTa~\cite{DBLP:journals/corr/abs-1907-11692} used strategies such as a larger corpus, a dynamic mask mechanism, and only applying MLM tasks for pre-training. ALBERT~\cite{DBLP:conf/iclr/LanCGGSS20} reduced BERT size by sharing layers parameters and compressing word embeds. These works have achieved very successful results.

However, unlike English, Chinese lacks explicit word boundary markers such as spaces between words, posing challenges for PLM-based sequence labeling tasks such as CWS, POS and NER~\cite{cui2021pre,wei2019nezha,lebert,babert}. Recent work explores methods to incorporate boundary information into Chinese PLMs. For MLM tasks, ERNIE-baidu~\cite{sun2019ernie} and BERT-wwm~\cite{cui2021pre} apply three different masking granularities — tokens, entities, and phrases, allowing the model to learn coarse-grained boundary information at the word and phrase levels rather than just at the character level. ERNIE-Gram~\cite{xiao-etal-2021-ernie} detects entities and phrases through statistical algorithms. 

In the unsupervised approach, BABERT leverages a large-scale corpus to extract a substantial amount of statistical boundary information~\cite{babert}. Building upon this unsupervised boundary information, a specific unsupervised boundary-aware learning objective is designed. While these methods successfully introduce boundary information, our paper's focus lies in combining the strengths of unsupervised statistical information with supervised high-quality information. We aim to leverage the advantages of both approaches to enhance the overall performance of the model.

\section{Conclusion}
This paper introduces Semi-BABERT, a model specifically designed for Chinese sequence labeling tasks. Semi-BABERT incorporates lexicon-based high-quality boundary information into BABERT through a span-based boundary recognition pre-training task. Experimental results on 13 sequence labeling datasets, including tasks such as CWS, POS, and NER, demonstrate that Semi-BABERT exhibits stronger boundary awareness than other PLMs like BABERT. 
Furthermore, Semi-BABERT demonstrates broad effectiveness across various NLP tasks. Additionally, we propose the Boundary Information Metric (BIM), which accurately quantifies the boundary encoding potential of Chinese PLMs without task-specific fine-tuning.

\section{Acknowledgement}
We sincerely thank the reviewers for their invaluable feedback, which significantly improved the quality of this work.
This work is supported by the National Natural Science Foundation of China (NSFC) Grant Nos.62336008 and Nos.62176180.

\nocite{*}

\bibliographystyle{lrec-coling2024-natbib}
\bibliography{lrec-coling2024-example}

\bibliographystylelanguageresource{lrec-coling2024-natbib}
\bibliographylanguageresource{languageresource}
\end{CJK}

\end{document}